# A Study of Deep Feature Fusion based Methods for Classifying Multi-lead ECG

**Bin Chen[1,2], Wei Guo[1], Bin Li[1], Rober K. F. Teng[1], Mingjun Dai[1,2], Jianping Luo[1], Hui Wang[1]**
1. College of Information Engineering, Shenzhen University, Shenzhen 518060, China.
2. The State Key Laboratory of Integrated Services Networks, Xidian University, Xian 710126, China.

Corresponding author: Bin Chen (e-mail: bchen@szu.edu.cn).

The research was jointly supported by research grant from Natural Science Foundation of China (61575126), Basic Research foundation of Shenzhen City (JCYJ20170818091801577, CYJ20160331114526190, JCYJ20170302145554126), the Key Project of Department of Education of Guangdong Province (2015KTSCX121)，xiniuniao from Tencent, and Natural Science Foundation of Shenzhen University (00002501).

**ABSTRACT** An automatic classification method has been studied to effectively detect and recognize Electrocardiogram (ECG). Based on the synchronizing and orthogonal relationships of multiple leads, we propose a Multi-branch Convolution and Residual Network (MBCRNet) with three kinds of feature fusion methods for automatic detection of normal and abnormal ECG signals. Experiments are conducted on the Chinese Cardiovascular Disease Database (CCDD). Through 10-fold cross-validation, we achieve an average accuracy of 87.04% and a sensitivity of 89.93%, which outperforms previous methods under the same database. It is also shown that the multi-lead feature fusion network can improve the classification accuracy over the network only with the single lead features.

## I. INTRODUCTION

Currently, the analysis and diagnosis of electrocardiogram are mainly performed by doctors who have experience and knowledge in the corresponding field and have accumulated years of experience. A lack of experienced doctors is very likely to miss the patient's best time for treatment. Therefore, it is very important to study an automatic diagnostic technique for ECG signals.

Recently, deep learning has made breakthroughs in the fields of image and natural language processing. It has been widely applied to various fields for its powerful ability to fit and express data. In the medical field of diagnosis of pneumonia [1], prediction of heart attack [2], and prediction of autism [3] and etc., deep learning has surpassed doctors. Deep neural network training requires a lot of data [4-6]. In terms of cardiovascular diseases, due to the complexity and variability of cardiovascular disease conditions, the requirement of data volume is even higher.

There are four internationally authoritative databases, which are MIT-BIH ECG database, AHA arrhythmia ECG database [7], CSE ECG database [8] and ST-T ECG database [9]. The MIT-BIH ECG database is currently the most widely used database and consists of many sub-databases. Each sub-database contains a certain type of ECG records, of which the most used is the MIT-BIT arrhythmia database and the QT database [10-16]. The AHA arrhythmia database is used to evaluate ventricular arrhythmia detectors. The CSE ECG database is used to evaluate the performance of the ECG auto-analyzer. The ST-T ECG database is a database for evaluating the performance of ST-segment and T-wave detection algorithms. Table I compares these five ECG databases, the MIT-BIH arrhythmia database has 48 ECG records, The QT database has 105 ECG records, the AHA arrhythmia database has 155 ECG records, and the ST-T ECG database has 90 ECG records. In the first four databases, each ECG record is composed by 2 leads. The CSE ECG database has 1,000 ECG records, and each ECG record is composed by 12 leads.

Table I. Comparison of five ECG databases

| Database | MIT-BIH AR | QT | AHA | ST-T | CSE |
|---|---|---|---|---|---|
| The number of leads | 2 | 2 | 2 | 2 | 12 |
| ECG record number | 48 | 105 | 155 | 90 | 1000 |

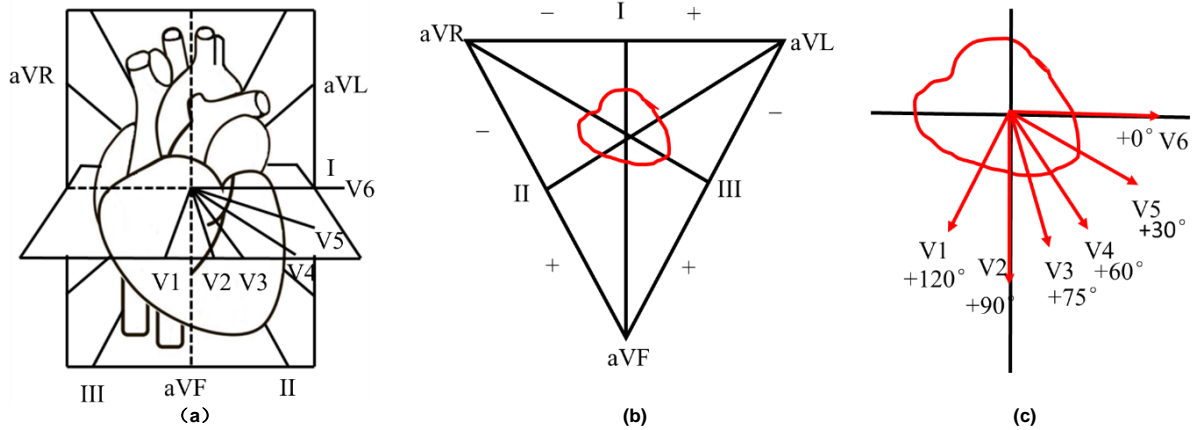

**Figure 1:** (a) The 12 lead positions of the conventional ECG are shown above. (b) The six leads of I, II, III, aVR, aVL, and aVF are called "frontal planes," and can only respond to upper, lower, left, and right plane electrical activity. (c) The six leads V1, V2, V3, V4, V5, and V6 are called "horizontal planes" and can only reflect the front, back, left, and right planes of this plane.

Currently the method of 12-lead simultaneous recording electrocardiographs is very popular in clinical practice. The position of 12 leads is shown in Figure 1. The superiority of the 12-lead simultaneous recording electrocardiograph lies in: 1) It can record the ECG signal of the same cardiac cycle on 12 leads at the same time. The identification and localization of single-source or multi-source premature beats, the classification of arrhythmia, and the diagnosis of indoor conduction block are superior to other electrocardiographs. 2) It greatly improves the accuracy of all measurements and reduces the variability of ECG measurement at present. 3) It can promote the establishment of basic measurement parameters such as P, QRS, T wave duration, and PR, Q-T interval and so on [18].

Deep-learning training requires a lot of data [4-6]. To achieve automatic classification of ECG and good generalization, the amount of data in above ECG databases is far from enough. Zhang J et al. established the 12-lead ECG database of the Chinese Cardiovascular Disease Database (CCDD) in 2010[19]. There are more than 190,000 12-lead ECG records in the CCDD database, each with at least one label. Jin L et al. proposed an LCNN model for classifying multi-lead ECG data in CCDD [20]. This model uses 8 leads. Each lead passes through three convolution layers, then goes through the fully connected layer. The model achieves 83.66% accuracy in automatically determining normal and abnormal samples. Based on this, in 2017, the team integrated two LCNNs and four rule-reasoning classifiers [21] to achieve 86.22% accuracy.

Recently, the residual network (ResNet) shows accuracy gains comparing with the traditional convolutional neural networks [22]. It has achieved great success in image and voice tasks and is another milestone in deep learning. In 2017, Rajpurkar P et al. [17] collected 64,121 single-lead ECG records from 29,163 patients. With the database, a model based on the ResNet was proposed and high classification accuracy is achieved. However, this databased is not available in the public. In this work, we will try to apply ResNet in multi-lead simultaneous recording electrocardiograph for improving the classification accuracy.

A "Multi-branch Convolution and Residual Network" (MBCRNet) is proposed and three multi-lead feature fusion methods are studied.

## II. Network Model

### A. The combination of multi-branch convolutional neural network and ResNet

In a convolutional neural network (CNN), the deeper the network, the more powerful it is [23]. However, increasing the depth of the network will cause the gradient to disappear or explode [24]. This makes it difficult to train the entire network. In 2016, He K et al. [22] proposed ResNet to solve this problem, which makes it possible for deep networks to continue being trained and get good results. In this work, we proposes a model to combine double-branch convolutional neural network [25] and ResNet (DBCRN) for abstracting features from single-lead ECG signal, which is shown in Figure 2:

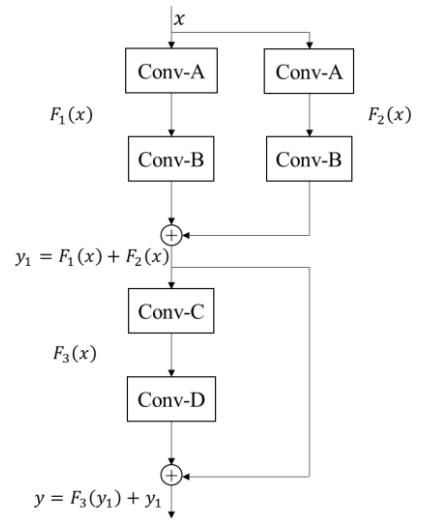

**Figure 2.** "DBCRN Block" is a combination of multi-branch convolutional neural networks and residual networks.

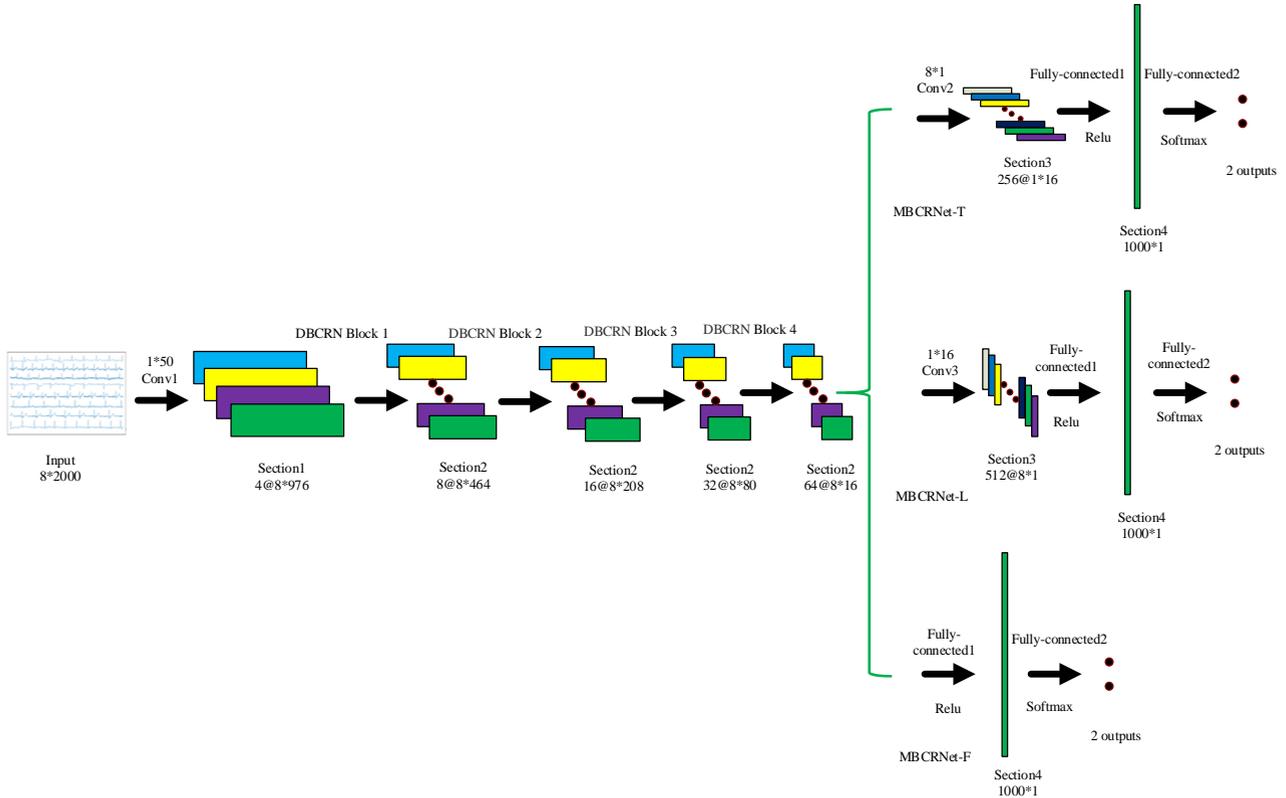

**Figure 3.** For the synchronization and orthogonality of multi-lead signals, three methods are proposed to fuse the extracted lead features.

In Fig. 2, signal *x* is a input into two convolutional neural network branches to reduce the dimension. Each branch has two convolution layers. The hyperparameters of the same layer, such as the size, depth, and step size of the convolution kernel, are the same. The features are extracted through two branches and then added. The extracted features have the same degree of abstraction. The summation may produce mutually reinforcing effects on the extracted features. In the experiment, we have found that more than two branches can only provide marginal performance improvement. Considering the efficiency, two branches are chosen in this work. It is noted that each "Conv" layer in the structure corresponds to sequence Conv-ReLU-BN. The residual network is realized by adding the forward neural network with "shortcut connection". Shortcut connections are those skipping one or more layers. Shortcut connections simply perform identity mapping. Their outputs are added to the outputs of the stacked layers. This structure in Figure 2 is named "DBCRN Block".

**B. Proposed Model**

The twelve leads in the clinical ECG are I, II, III, aVR, aVL, aVF, V1, V2, V3, V4, V5, V6 in synchronization. Among the leads, II, III, V1, V2, V3, V4, V5 and V6 leads are orthogonal. Remaining 4 leads can be linearly deduced from 8 basic leads [20]. In this article, we choose eight orthogonal leads for the feature fusion.

Considering the synchronization and orthogonality of multiple leads, we propose three multi-branch convolution and residual network (MBCRNet) frameworks which are shown in Fig. 3. In all frameworks, input signal is a two-dimensional array of $8 \times 2000$. Each line represents a lead. Each lead takes 2000 samples. DBCRN Block 1, DBCRN Block 2, DBCRN Block 3, and DBCRN Block 4 use the network structure as shown in Figure 2. In the four blocks, their network structures, convolution kernel size, and step sizes are the same. The only difference is the depth of the convolution kernel. The depths of the convolution kernels of DBCRN Block 1, DBCRN Block 2, DBCRN Block 3, and DBCRN Block 4 are 8, 16, 32 and 64 respectively. The input signal first passes a $1 \times 50$ convolution, and then passes through four DBCRN Blocks to extract features of each lead. The numbers before and after "@" refer to the number and dimension of the feature map in the corresponding layer.

**C. Feature Fusion**

Considering relationship between multiple leads, we propose three feature fusion methods. In the first one, we consider synchronization relationships among 8 leads and name it as MBCRNet-T. The extracted lead features are merged through an $8 \times 1$ convolution, which combine the characteristics of each lead in the same time interval. Finally, outputs of the convolution networks are sent to the full connection layer for classification. In the second method, we consider orthogonality relationships among 8 leads and name it as MBCRNet-L. Each extracted lead feature is fused by a 1×16 convolution before it is sent to a full connection network. In the third method, we let the full connection layer to find the relationship among lead features and to make decisions. It is named as MBCRNet-F. The specific network configuration is shown in Table II.

Table II. The network structure used for feature fusion in the CCDD database through three different methods. Note that each "Conv" layer corresponds the sequence Conv- BN-ReLU. Due to space, Dropout layer is not shown.

| Layers | Type | Output Size | MBCRNet-T | MBCRNet-L | MBCRNet-F |
|---|---|---|---|---|---|
| Conv1 | Convolution | 8 x 976 | 1 × 50 Conv stride 1 × 2 | | |
| DBCRN Block 1 | Double-branch Conv | 8 x 464 | $\begin{bmatrix} 1 \times 50 \text{ Conv stride } 1 \times 2, 1 \times 50 \text{ Conv stride } 1 \times 2 \\ 1 \times 50 \text{ Conv stride } 1 \times 1, 1 \times 50 \text{ Conv stride } 1 \times 1 \end{bmatrix}$ | | |
| | ResNet | 8 x 464 | $\begin{bmatrix} 1 \times 50 \text{ Conv} \\ 1 \times 50 \text{ Conv} \end{bmatrix}$ | | |
| DBCRN Block 2 | Double-branch Conv | 8 x 208 | $\begin{bmatrix} 1 \times 50 \text{ Conv stride } 1 \times 2, 1 \times 50 \text{ Conv stride } 1 \times 2 \\ 1 \times 50 \text{ Conv stride } 1 \times 1, 1 \times 50 \text{ Conv stride } 1 \times 1 \end{bmatrix}$ | | |
| | ResNet | 8 x 208 | $\begin{bmatrix} 1 \times 50 \text{ Conv} \\ 1 \times 50 \text{ Conv} \end{bmatrix}$ | | |
| DBCRN Block 3 | Double-branch Conv | 8 x 80 | $\begin{bmatrix} 1 \times 50 \text{ Conv stride } 1 \times 2, 1 \times 50 \text{ Conv stride } 1 \times 2 \\ 1 \times 50 \text{ Conv stride } 1 \times 1, 1 \times 50 \text{ Conv stride } 1 \times 1 \end{bmatrix}$ | | |
| | ResNet | 8 x 80 | $\begin{bmatrix} 1 \times 50 \text{ Conv} \\ 1 \times 50 \text{ Conv} \end{bmatrix}$ | | |
| DBCRN Block 4 | Double-branch Conv | 8 x 16 | $\begin{bmatrix} 1 \times 50 \text{ Conv stride } 1 \times 2, 1 \times 50 \text{ Conv stride } 1 \times 2 \\ 1 \times 50 \text{ Conv stride } 1 \times 1, 1 \times 50 \text{ Conv stride } 1 \times 1 \end{bmatrix}$ | | |
| | ResNet | 8 x 16 | $\begin{bmatrix} 1 \times 50 \text{ Conv} \\ 1 \times 50 \text{ Conv} \end{bmatrix}$ | | |
| Conv2 | Convolution | 1 x 16 | 8 × 1 Conv | | |
| Conv3 | Convolution | 8 x 1 | | 1 × 16 Conv | |
| Fully-connected1 | Fully-connected | 1000 x 1 | 1000D fully-connected, Relu | | |
| Fully-connected2 | Fully-connected | 2 x 1 | 2D fully-connected, softmax | | |

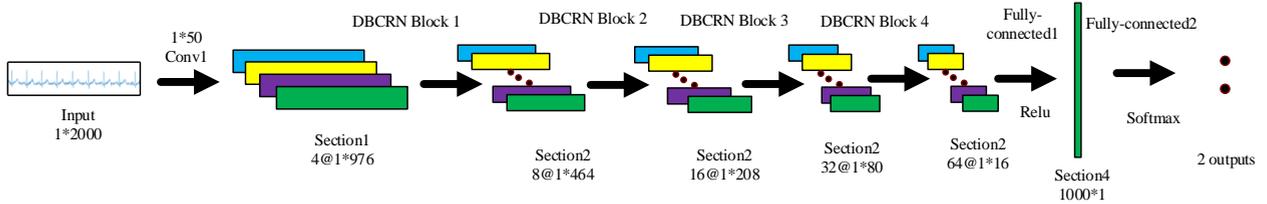

**Figure 4.** The network structure used to train single lead ECG.

### III. Experimental Setup

#### A. Data
Chinese Cardiovascular Disease Database (CCDD) has over 190,000 ECG records. The ECG data is sampled at a frequency of 500 Hz and is collected from 12 leads. The duration of each ECG record is about 10s. In this work, we use the same method to mark normal and abnormal ECG as in [20]. Records labeled "normal electrocardiogram" and "normal sinus rhythm" are considered as normal (labeled as "0"). The others are considered as abnormal (labeled "1"). The data preprocessing steps are as follows: (1) Invalid ECG and ECG with recording time less than 8 seconds are removed. 90,804 normal samples and 72,083 abnormal samples are obtained. 70,000 normal samples and 70,000 abnormal samples are randomly selected for the second step. (2) The sampling rate of raw data are reduced from 500 Hz to 250 Hz. Only 8 leads (II, III, V1, V2, V3, V4, V5, V6) in each sample are used. For each sample, 8 seconds records or 2000 sampling values for each lead are extracted. (3) We do 10 fold cross validation in the experiment. The proportion of positive and negative samples in the training set and test set is 1:1. The samples are divided into 10 equal size sets. One of them is selected as a test set, and the other 9 sets are used for training.

#### B. Training and Testing
The experiment is conducted on a workstation with one Intel Xeon 3.60 GHz (i7-6850k) processor and 32 GB RAM specification. For the classification of multi-lead ECG signals, the 8 lead data of extracted II, III, V1, V2, V3, V4, V5, and V6 are organized into an 8×2000 two-dimensional array and input into the network. In order to compare the classification accuracy between single-lead data and multi-lead data of ECG, 8 single leads are separately trained and classified in the experiment, and 10-fold cross validation was performed. The network structure used is shown in Figure 4, which is the same as the network framework of MBCRNet-F with only one lead. All results are discussed as following.

Table III. Accuracy and sensitivity of 10-fold cross-validation test set for 8 leads

| 8 Leads | MBCRNet-T | | MBCRNet-L | | MBCRNet-F | |
|---|---|---|---|---|---|---|
| | ACC | Se | ACC | Se | ACC | Se |
| Fold-1 | 86.13% | 89.33% | 86.64% | 89.67% | 85.91% | 87.97% |
| Fold-2 | 87.42% | 90.53% | 87.41% | 90.96% | 87.13% | 88.67% |
| Fold-3 | 86.61% | 89.37% | 87.31% | 90.57% | 86.99% | 88.90% |
| Fold-4 | 86.29% | 89.84% | 87.09% | 90.63% | 86.71% | 87.24% |
| Fold-5 | 86.68% | 88.77% | 86.79% | 90.03% | 86.88% | 90.30% |
| Fold-6 | 86.40% | 88.76% | 86.86% | 89.74% | 86.42% | 88.01% |
| Fold-7 | 86.78% | 88.74% | 86.89% | 88.27% | 86.47% | 85.67% |
| Fold-8 | 86.34% | 88.43% | 86.94% | 90.33% | 86.67% | 89.13% |
| Fold-9 | 86.45% | 86.63% | 87.40% | 89.29% | 86.22% | 88.29% |
| Fold-10 | 86.76% | 89.23% | 87.04% | 89.84% | 86.52% | 89.36% |
| Average | 86.59% | 88.96% | 87.04% | 89.93% | 86.59% | 88.35% |

C. Results

This paper uses two performance indicators, Accuracy (ACC) and Sensitivity (Se), to evaluate the capabilities of the model. 10-fold cross validation results for 8-lead ECG data are shown in Table III. The average accuracy values of MBCRNet-T, MBCRNet-L, and MBCRNet-F methods are 86.59%, 87.04%, and 86.59% respectively. The average sensitivity values of MBCRNet-T, MBCRNet-L, and MBCRNet-F methods are 88.96%, 89.93% and 88.35% respectively. It is shown that the average accuracy and average sensitivity values under the MBCRNet-L method are higher than those under MBCRNet-T and MBCRNet-F methods. In the MBCRNet-L method, the feature of each lead is extracted and seperated before being input to full connected network for classification. On the other side, in MBCRNet-T method, the features of different lead in the same time interval are fused before being input into the full connected network. In MBCRNet-F method, features of a lead in different time interval are mixed among different leads before classification. Therefore, we conclude that the feature of whole lead should not be separated and mixed in time interval from others.

Table IV. Comparison of experimental results

| | Model | ACC | Se |
|---|---|---|---|
| Jin L [20] | LCNN | 83.66% | ~ |
| Jin L [21] | LCNN+ Rule Inference | 86.22% | 86.46% |
| Our method | MBCRNet-L | 87.04% | 89.93% |

Table IV compares our MBCRNet-L method with other methods using the CCDD database. Authors in [20] used the LCNN model to obtain 83.66% of the normal and abnormal classification accuracy. In [21], authors combine the LCNN model with four rule inferences and obtained 86.22% of the classification accuracy. The end-to-end MBCRNet-L model based on the residual network achieve 87.06% accuracy and 89.93% sensitivity. The accuracy and sensitivity of MBCRNet-L are better than the two previous methods.

Figure 5 compares the accuracy of single lead classification and multi-lead classification. For the normal and abnormal ECG classifications, the average accuracy of lead III is 78.96% which is lower than the average accuracy values of other 7 leads. The average accuracy of V5 lead is 85.19%, which is the highest among average accuracy values of single leads. The average accuracy of 8 single leads is 81.66%. The average accuracy values under all multi-lead classification methods are higher than the highest single lead classification average accuracy value of V5. Therefore, we conclude that the multi-lead feature fusion can improve the classification accuracy for ECG signals, which is consistent with clinical experience [18].

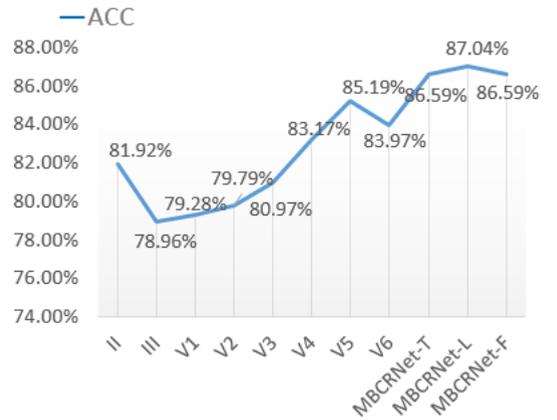

Figure 5. Comparison of average accuracy of single-lead and multi-lead classifications.

IV. Conclusion

In this paper, we study end-to-end auto-detection deep architecture for multi-lead synchronous ECG signals. This deep architecture uses multi-branch convolution and ResNet to capture characteristics of different lead ECG signals. The extracted features are fused by convolution, and eventually map to different classes. In the experiment, the classification of normal and abnormal electrocardiograms for multi-lead ECG signals has yielded an average accuracy of 87.04% and a sensitivity of 89.93% by 10-fold cross validation, which verifies that the multiple-lead feature fusion can improve the classification accuracy comparing with the network only using single lead features. It is also proved that the proposed model based on ResNet is superior to other methods with the CCDD database. Therefore, the method we proposed in this work can be used as a competitive tool for feature learning and classification for multi-lead ECG signal classification problems.